# Empirical Probabilities in Monadic Deductive Databases

Raymond T. Ng and V.S. Subrahmanian
Department of Computer Science
A. V. Williams Building
University of Maryland
College Park, Maryland 20742, U.S.A.

## Abstract

We address the problem of supporting empirical probabilities in monadic logic databases. Though the semantics of multivalued logic programs has been studied extensively, the treatment of probabilities as results of statistical findings has not been studied in logic programming/deductive databases. We develop a model-theoretic characterization of logic databases that facilitates such a treatment. We present an algorithm for checking consistency of such databases and prove its total correctness. We develop a sound and complete query processing procedure for handling queries to such databases.

## 1 Introduction

During the past decade, there has been growing interest in reasoning with uncertainty in logic programming, deductive databases and expert systems. The semantics of logic programs based on uncertainty has been studied by van Emden [20], Blair and Subrahmanian [5], Dubois et al. [6], Kifer et. al. [11] and Fitting [8]. All these studies were based upon non-probabilistic modes of uncertainty. The first semantical characterizations of probabilistic logic programming were those of Ng and Subrahmanian [13, 14]. These characterizations were based on *subjective* probabilities using a Kripke-style possible-worlds semantics along the lines of the work by Nilsson [16].

While subjective probabilities view probabilities as degrees of belief, *empirical* probabilities represent objective, statistical truths about the world/population. Typically, empirical probabilities are obtained from samples of individuals in the population. These samples can then used to inductively infer probabilities about the entire population. For example, if 500 monk seals are physically caught (in some unbiased way), these 500 monk seals constitute the sample in question. Inductive reasoning allows us to draw conclusions about a specific monk seal that is not in the sample.

The main aim of this paper is to capture this kind of empirical reasoning within the framework of deductive databases. The knowledge about specific individuals in the sample is two-valued true/false knowledge. This knowledge is represented as a set of ordinary first order logic clauses called the *context*. The probabilities derived from the sample are expressed as *conditional probability* statements and constitute the empirical component of our knowledge. These probabilities may then be used to *induce* information about objects not in the context (sample), but in the population. As we are studying only individuals and their properties, rather than the relationships between individuals, these properties are represented using unary (or monadic) predicates.

The first contribution of this paper is the development of a formal model-theoretic basis for such databases. This task is complicated by the fact that Herbrand interpretations are not rich enough to capture the intuitions behind empirical probabilities (cf. Example 2). The second contribution is the development of sound and complete algorithms for determining the consistency of such databases. These algorithms are based on constraint satisfaction techniques, and hence, may be implemented using standard mixed integer linear programming algorithms. The third contribution is a procedure to answer queries to such databases. This procedure first tries to deduce a definite answer from the context, and if that is not possible, it tries to induce answers from the empirical component. This procedure is sound and complete.

## 2 Empirical Programs

Let $\mathcal{L}$ be a language generated by finitely many *unary* predicate symbols, but no function symbols. In this paper, we are interested in the properties of individuals. Hence, we regard a predicate symbol as a representation of a property that individuals in the domain of discourse may or may not have. Essentially, $\mathcal{L}$ allows us to express properties about individuals, even though it does not support properties or relations among a group of individuals.

**Definition 1** A *context* $C$ is a finite set of clauses of either one of the following two forms:
i) $K_0 \leftarrow K_1 \wedge \ldots \wedge K_n$, where for all $0 \leq i \leq n$, $K_i$ is a *ground* literal, or
ii) $L_0(X) \leftarrow L_1(X) \wedge \ldots \wedge L_n(X)$, where for all $0 \leq i \leq n$, $L_i \equiv A_i$ or $L_i \equiv \neg A_i$ for some predicate symbol $A_i$ in $\mathcal{L}$.                                    □

A non-ground clause in a context is implicitly universally quantified at the front of the clause. In other words, the clause $L_0(X) \leftarrow L_1(X) \wedge \ldots \wedge L_n(X)$ says that for all individuals $X$, if $X$ has properties $L_1, \ldots, L_n$, then $X$ also has the property $L_0$. Under this interpretation, we disallow clauses such as $A(X) \leftarrow B(Y)$ (for some variable symbol $Y \in \mathcal{L}$) and $A(X) \leftarrow B(c)$ (for some constant $c \in \mathcal{L}$) from appearing in a context, as we find such clauses unintuitive for our purpose. An additional point worth mentioning is that all negations are interpreted classically. Hence, these are not treated in terms of the Closed World Assumption or other such non-monotonic rules of inference.

We now proceed to define *empirical clauses*. A feature of these clauses is that "complex properties" are embodied syntactically by the notion defined below.

**Definition 2** A *complex* predicate symbol $F$ is an arbitrary disjunction and/or conjunction of positive and/or negated predicate symbols in $\mathcal{L}$.        □

As an example, if $A$, $B$ and $C$ are predicate symbols, then $A$, $(A \vee B)$, $((A \wedge B) \vee C)$ and $((\neg A) \wedge B \wedge (\neg C))$ are some complex predicate symbols that can be generated from these three predicate symbols. Furthermore, given that $X$ is a variable symbol in $\mathcal{L}$, we use the notations $(F_1 \wedge F_2)(X)$ and $F_1(X) \wedge F_2(X)$ interchangeably, where $F_1, F_2$ are two complex predicate symbols. Similarly, $(F_1 \vee F_2)(X)$ and $F_1(X) \vee F_2(X)$ are the same. Complex predicate symbols may be viewed as propositional formulas with the predicate symbols of $\mathcal{L}$ being considered as propositions rather than as unary predicates.

**Definition 3** 1) An *empirical* clause is of the form:
$[c_1, c_2] \; F(X) \quad \leftarrow \quad F_1(X) \wedge \ldots \wedge F_n(X)$ where $F, F_1, \ldots, F_n$ are complex predicate symbols, $X$ is a variable, and $c_1, c_2$ are real numbers in $[0,1]$ such that $c_1 \leq c_2$ and $[c_1, c_2]$ is equal to neither $[1,1]$ nor $[0,0]$.
2) An *empirical program* $P = \langle C, E \rangle$ consists of a context $C$ and a finite set $E$ of empirical clauses.        □

The intended meaning of an empirical clause is: "given that an arbitrary individual $X$ has (complex) properties $F_1, \ldots, F_n$, the conditional probability of $X$ having property $F$ falls within the range $[c_1, c_2]$." From an empirical point of view, it states that: "amongst all individuals that have properties $F_1, \ldots, F_n$, between $c_1\%$ to $c_2\%$ of them have property $F$." Note that an empirical clause is *not* meant to include an implicit universal quantifier on $X$. In other words, it is not a statement about a universally quantified variable $X$; rather, it is a statement about a "generic" individual $X$ in the domain of discourse.

In classical logic, the clause $A(X) \leftarrow (B(X) \vee C(X))$ can be represented by the two clauses: $A(X) \leftarrow B(X)$ and $A(X) \leftarrow C(X)$. Similarly, $A(X) \wedge B(X) \leftarrow C(X)$ is equivalent to the conjunction of the clauses $A(X) \leftarrow C(X)$ and $B(X) \leftarrow C(X)$. One may wonder whether such an equivalence is true for empirical clauses; if so, complex predicate symbols can be dispensed with. In [15], we show that this is not always possible, and thus, the presence of complex predicate symbols makes empirical clauses more expressive than those without complex symbols.

**Example 1** Consider the empirical program $\langle C, E \rangle$ where $E$ consists of the following clauses:

$$[0, 0.1] \; Grey(X) \quad \leftarrow \quad Royal\_elephant(X)$$
$$[0.9, 0.95] \; Grey(X) \quad \leftarrow \quad Elephant(X)$$
$$[0.1, 0.2] \; Elephant(X) \quad \leftarrow \quad Grey(X)$$

The first clause states that at most 10% of "Royal_elephants" are "Grey." The last two clauses describe the relative percentages of "Elephants" and "Grey" individuals. In particular, they state that between 90% and 95% of all "Elephants" are "Grey," whereas only between 10% and 20% of "Grey" individuals are "Elephants." Finally, let the context $C$ consist of the following clauses:

$$Elephant(X) \quad \leftarrow \quad Royal\_elephant(X)$$
$$Grey(X) \quad \leftarrow \quad \neg White(X)$$
$$Royal\_elephant(clyde) \quad \leftarrow$$
$$Elephant(jill) \quad \leftarrow$$

The second clause states that for this particular context, every individual is either "Grey" or "White."  □

## 3    Model Theoretic Semantics

### 3.1    Interpretations and Models

For our language $\mathcal{L}$, the notion of an interpretation is almost identical to the usual one for first-order languages – except for two differences. The first one is that interpretations $I$ are assumed to map different constant symbols in $\mathcal{L}$ to distinct elements in the domain of $I$. The second difference is that the domain of $I$ is assumed to be finite. There are two reasons for this assumption. The first one is that empirical programs are intended to reflect findings from statistical samples which are always finite. The other reason, which is more technical, will be apparent in a later definition on *satisfaction* (cf. Definition 4).

In conventional logic programming, it suffices to use Herbrand interpretations. However, the following example shows that when dealing with empirical probabilities, using only Herbrand interpretations and universes may not make sense at all.

**Example 2** Consider the following empirical program:

$$C: \qquad Monk\_seal(joe) \quad \leftarrow$$
$$E: \quad [0.4, 0.45] \, Female(X) \quad \leftarrow \quad Monk\_seal(X)$$

The empirical clause states that between 40% and 45% of all "Monk_seals" are "Females." The context part consists only of the single fact that *joe* is a "Monk_seal." But note that the Herbrand universe is the singleton $\{joe\}$. Since there is only 1 member in this set, what then is the meaning of the statement that between 40% and 45% of all "Monk_seals" are "Females"? In particular, there are two possible Herbrand interpretations that will satisfy the second clause, namely $\{Monk\_seal(joe)\}$ and $\{Monk\_seal(joe), Female(joe)\}$. In both cases, $Female(joe)$ is either true or false. With only these two interpretations, it is unclear how they can "satisfy" the first clause in an empirical sense.                  □

As the above example shows, considering Herbrand interpretations alone may fail to capture the empirical nature of probabilistic reasoning. Hence, in order to discuss arbitrary domains, we also consider non-Herbrand interpretations. We will define shortly when an interpretation satisfies an empirical program. In the sequel, given a predicate symbol $A$ and an interpretation $I$ with domain $D$, we use the notation $A_I$ to denote the set $\{d \in D \mid \phi_A(d) = true\ \}$, where $I$ maps the predicate symbol $A$ to the mapping $\phi_A$. We extend this notation to complex predicate symbols: $(F_1 \wedge F_2)_I$ to mean $(F_{1I} \cap F_{2I})$, and $(F_1 \vee F_2)_I$ to mean $(F_{1I} \cup F_{2I})$, where $F_1, F_2$ are two complex predicate symbols. We also use the notation $\|F_1 \wedge \ldots \wedge F_n\|_I$ to denote the cardinality of the set $F_{1I} \cap \ldots \cap F_{nI}$. Intuitively, $\|F_1 \wedge \ldots \wedge F_n\|_I$ denotes "the number of individuals in the domain of $I$ that have properties $F_1, \ldots, F_n$." Before we define the notion of satisfaction, recall that an empirical program consists of a context $C$ and a set $E$ of empirical clauses. Given the nature of $C$, we say that $I$ satisfies $C$ iff $I$ satisfies every clause in $C$ in the usual sense, for first-order languages. Thus, we only need to define the condition for $I$ to satisfy the set $E$ of empirical clauses.

**Definition 4** Let $(C, E)$ be an empirical program and $I$ be an interpretation.
1) We say that $I$ satisfies the empirical clause $[c_1, c_2]\ F(X) \leftarrow F_1(X) \wedge \ldots \wedge F_n(X)$ iff:
i) $\|F_1 \wedge \ldots \wedge F_n\|_I = 0$, or

ii) whenever $\|F_1 \wedge \ldots \wedge F_n\|_I > 0$, $c_1 \leq \frac{\|F \wedge F_1 \wedge \ldots \wedge F_n\|_I}{\|F_1 \wedge \ldots \wedge F_n\|_I} \leq c_2$.

2) We say that $I$ satisfies $E$ iff $I$ satisfies every empirical clause in $E$. Finally, $I$ satisfies the empirical program $(C, E)$ iff $I$ satisfies both $C$ and $E$.                 □

**Example 3** Consider the empirical program in Example 2. Let $I$ be the interpretation defined as follows. Firstly, the domain of $I$ is the set $\{d_1, \ldots, d_{20}\}$, for some individuals $d_1, \ldots, d_{20}$. Secondly, $I$ assigns to the constant symbol *joe*, the element $d_5$. Furthermore, $I$ assigns to the predicate symbol $Monk\_seal$, the mapping $f_1$ such that $f_1(d_i) = true$ for all $1 \leq i \leq 10$, and $f_1(d_i) = false$ for all $11 \leq i \leq 20$. Finally, $I$ assigns to the predicate symbol $Female$, the mapping $f_2$ such that $f_2(d_i) = true$ for all $1 \leq i \leq 4$, and $f_1(d_i) = false$ for all $5 \leq i \leq 20$. Then, it is straightforward to check that $I$ is a model of the program.                 □

Recall that the intuitive meaning of an empirical clause $[c_1, c_2]\ F(X) \leftarrow F_1(X) \wedge \ldots \wedge F_n(X)$ is: "given that an arbitrary individual $X$ has (complex) properties $F_1, \ldots, F_n$, the conditional probability that $X$ also has property $F$ falls within the range $[c_1, c_2]$." According to probability theory, the statement can be represented as: $c_1 \leq Prob(F|F_1 \cap \ldots \cap F_n) = \frac{Prob(F \cap F_1 \cap \ldots \cap F_n)}{Prob(F_1 \cap \ldots \cap F_n)} \leq c_2$, where $F, F_1, \ldots, F_n$ now represent the appropriate events and assuming that $Prob(F_1 \cap \ldots \cap F_n) \neq 0$. Furthermore, when probabilities are interpreted empirically over statistical samples, we have the following: $\frac{Prob(F \cap F_1 \cap \ldots \cap F_n)}{Prob(F_1 \cap \ldots \cap F_n)} = \frac{Num(F \cap F_1 \cap \ldots \cap F_n)}{Num(F_1 \cap \ldots \cap F_n)}$, where $Num(S)$ denotes the cardinality of the set $S$. This is the intuition behind Definition 4 for an interpretation to satisfy an empirical clause.

Finally, recall that an interpretation $I$ for $\mathcal{L}$ has a finite domain. A technical reason for this restriction is that then the ratio $\frac{\|F \wedge F_1 \wedge \ldots \wedge F_n\|_I}{\|F_1 \wedge \ldots \wedge F_n\|_I}$ is always well-defined and is between 0 and 1, whenever $\|F_1 \wedge \ldots \wedge F_n\|_I > 0$.

### 3.2 Consistency of Empirical Programs with Empty Contexts

Let $\{A_1, \ldots, A_k\}$ be the set of all predicate symbols in $\mathcal{L}$. Furthermore, suppose that all possible subsets of $\{A_1, \ldots, A_k\}$ are enumerated in an arbitrary, but fixed way: $\mathcal{P}_1, \ldots, \mathcal{P}_{2^k}$. For all $1 \leq i \leq 2^k$, let $v_i$ be an integer variable that intuitively denotes the number of individuals in an underlying domain that have the complex property: $\bigwedge_{A \in \mathcal{P}_i} A \bigwedge_{A \notin \mathcal{P}_i} \neg A$. It is easy to see that the $\mathcal{P}_i$'s divide the set of all individuals into $2^k$ partitions. Thus, given the $v_i$'s, the number of individuals that have an arbitrary property $F$ can also be determined. This amounts to checking whether for all $1 \leq i \leq 2^k$, $F$ is true in each $\mathcal{P}_i$ – in the classical 2-valued Herbrand sense. We often abuse notation by regarding $F$ as a proposition and write $\mathcal{P}_i \models F$ whenever $F$ is true in $\mathcal{P}_i$. Then the number of individuals having property $F$ is the summation of the number of individuals in each $\mathcal{P}_i$ satisfying $F$, i.e. $\sum_{\mathcal{P}_i \models F; i=1,\ldots,2^k} v_i$. Throughout this paper, to simplify our notation, we often write the term $(\sum_{\mathcal{P}_i \models F; i=1,\ldots,2^k} v_i)$ as $(\sum_{\mathcal{P}_i \models F} v_i)$.

**Example 4** Consider the empirical clause $[0.4, 0.45] Female(X) \leftarrow Monk\_seal(X)$ again. The two predicate symbols give rise to the following four partitions:

| $Monk\_seal$ | $Female$ | partition | number |
|:---:|:---:|:---:|:---:|
| 1 | 1 | $\mathcal{P}_1$ | $v_1$ |
| 1 | 0 | $\mathcal{P}_2$ | $v_2$ |
| 0 | 1 | $\mathcal{P}_3$ | $v_3$ |
| 0 | 0 | $\mathcal{P}_4$ | $v_4$ |

The first partition corresponds to individuals having the properties "Monk_seal" and "Female," and so on. Now, since the property "Monk_seal" is true in the classical 2-valued Herbrand sense in $\mathcal{P}_1 \equiv \{Monk\_seal, Female\}$ and $\mathcal{P}_2 \equiv \{Monk\_seal\}$, the number of individuals having the property "Monk_seal" is given by $(v_1 + v_2)$.    □

**Definition 5** Let $E$ be a set of empirical clauses.
1) Let $Cl \equiv [c_1, c_2] \, F(X) \leftarrow F_1(X) \wedge \ldots \wedge F_n(X)$ be an empirical clause. The *constraint version* of $Cl$, denoted by $con(Cl)$, is the constraint:

$$c_1 * (\sum_{\mathcal{P}_i \models F_1 \wedge \ldots \wedge F_n} v_i) \leq (\sum_{\mathcal{P}_i \models F \wedge F_1 \wedge \ldots \wedge F_n} v_i) \leq$$
$$c_2 * (\sum_{\mathcal{P}_i \models F_1 \wedge \ldots \wedge F_n} v_i)$$

2) Let $con(E)$ denote the set $\{con(Cl) | Cl \in E\} \cup \{\sum_{i=1}^{2^k} v_i \geq 1\}$.    □

The reason why we study $con(E)$ is that we intend to check the consistency of $E$ based on the constraints in $con(E)$. More specifically, if $I$ is a model of $E$, and if $n_i$ denotes the number of individuals in the domain of $I$ corresponding to $\mathcal{P}_i$ for all $1 \leq i \leq 2^k$, then the assignment $v_i = n_i$ for all $1 \leq i \leq 2^k$ represents a solution to the constraints in $con(E)$. Furthermore, the constraint $\sum_{i=1}^{2^k} v_i \geq 1$ in $con(E)$ guarantees that if $I$ is model of $E$, the domain of $I$ must be non-empty. While the details of this discussion will be formalized shortly, first consider the following example.

**Example 5** Let $E$ consist of the single clause $[0.4, 0.45] Female(X) \leftarrow Monk\_seal(X)$. Then according to Example 4, $con(E)$ consists of the following two constraints:

$$0.4 * (v_1 + v_2) \leq v_1 \leq 0.45 * (v_1 + v_2)$$
$$v_1 + v_2 + v_3 + v_4 \geq 1$$

One integer solution to this set of constraints is $v_1 = 4$, $v_2 = 6$, $v_3 = 0$, $v_4 = 10$. Thus, for instance, the

interpretation considered in Example 3 satisfies these constraints.    □

**Definition 6** Let $I$ be an interpretation. Let $S_I$ be defined as follows: for all $1 \leq i \leq 2^k$, $S_I(\mathcal{P}_i) = \| \bigwedge_{A \in \mathcal{P}_i} A \, \bigwedge_{A \notin \mathcal{P}_i} \neg A \|_I$.    □

Intuitively, $S_I$ specifies the number of individuals within each partition $\mathcal{P}_i$ of the domain of $I$.

**Theorem 1** Let $I$ be an interpretation and $E$ be a set of empirical clauses. $I$ is a model of $E$ iff $S_I$ is a solution of $con(E)$.    □

The above theorem suggests a practical way to check the consistency of a set $E$ of empirical clauses. This amounts to checking whether $con(E)$ has an integer solution. As all the constraints are linear [1], checking whether $con(E)$ has a solution can be carried out by an integer linear programming algorithm. One advantage of viewing consistency checking as constraint satisfaction is that implementations of such algorithms are widely available, such as the LINDO package running on the IBM/PC.

### 3.3 Consistency of Empirical Programs

Given an empirical program $\langle C, E \rangle$, recall from Definition 4 that an interpretation $I$ is a model of the program iff $I$ satisfies both $C$ and $E$. Theorem 1 above suggests a way to check the consistency of the $E$ part of the program. Moreover, there are certainly many ways to check the consistency of the $C$ part, such as using the implementation described in [4]. Obviously, the problem is that the consistency of $C$ and $E$, when considered separately, does not guarantee the joint consistency of $\langle C, E \rangle$. One straightforward solution to this problem is to find a model for the $C$ part and then test for satisfaction of the $E$ part using Theorem 1. However, if $\langle C, E \rangle$ is jointly inconsistent, this strategy may not terminate, as $C$ may have infinitely many models. In the following, we present a terminating procedure that decides the consistency of an empirical program.

**Algorithm 1** Let the input be an empirical program $\langle C, E \rangle$.

1. Partition the clauses in $C$ into two sets $C_1, C_2$ such that $C_1$ consists of all non-ground clauses in $C$ (i.e. $C_1 = \{Cl \mid Cl \equiv L_0(X) \leftarrow L_1(X) \wedge \ldots \wedge L_n(X)$ is a clause in $C\}$), and $C_2 = C - C_1$.

2. By Definition 1, every clause in $C_2$ is ground. Find all Herbrand models of $C_2$ using techniques such as the one described in [4].

---

[1] Strictly speaking, our constraints are fractional constraints. However, for most applications, these fractional constraints can be easily translated to linear constraints. See [19] for a general treatment on a translation from fractional to linear constraints.

3. If no such Herbrand model can be found, declare that the program is inconsistent and halt.

4. Otherwise, let $\{M_1, \ldots, M_u\}$ be the set of Herbrand models of $C_2$ computed in Step 2. Set $j$ to 1.

5. If $j > u$, declare that the empirical program is inconsistent and halt.

6. Initialize the set $T$ to $con(E)$.

7. For each clause $Cl \equiv L_0(X) \leftarrow L_1(X) \wedge \ldots \wedge L_n(X)$ in $C_1$, add the constraint: $(\sum_{\mathcal{P}_i \models L_0 \wedge L_1 \wedge \ldots \wedge L_n} v_i = \sum_{\mathcal{P}_i \models L_1 \wedge \ldots \wedge L_n} v_i)$ to $T$.

8. For each predicate symbol $A$ appearing in $C_2$, do the following:

   - Compute the number $n_A$ which is the cardinality of the set $\{A(c) \mid A(c) \in M_j$ for some constant symbol $c\}$. Then add the constraint: $(\sum_{\mathcal{P}_i \models A} v_i \geq n_A)$ to $T$.

   - Similarly, compute the number $\bar{n}_A$ which is the cardinality of the set $\{\neg A(c) \mid A(c) \notin M_j$ for some constant symbol $c\}$. Then add the constraint: $(\sum_{\mathcal{P}_i \models \neg A} v_i \geq \bar{n}_A)$ to $T$.

9. Determine whether the constraints in $T$ have a solution. If they do, declare that the program is consistent and halt.

10. Otherwise, increment $j$ and go to Step 5.    □

In Step 1 of Algorithm 1, the context $C$ is partitioned into two sets: $C_1$ consisting of all non-ground clauses and $C_2$ consisting of all ground clauses in $C$. As the number of clauses in $C_2$ is finite and all clauses are ground, the computation in Step 2 always terminates with a finite set of Herbrand models If this set is empty, then $\langle C, E \rangle$ is inconsistent. Otherwise, for each Herbrand model $M_j$ computed, Step 8 adds constraints of the forms $(\sum_{\mathcal{P}_i \models A} v_i \geq n_A)$ and $(\sum_{\mathcal{P}_i \models \neg A} v_i \geq \bar{n}_A)$ to $T$ for each predicate symbol $A$. This is to guarantee that subsequently an interpretation that corresponds to a solution of the constraints in $T$, will be able to make all ground atoms $A(c)$ true iff $A(c) \in M_j$. Finally, for Step 9, any standard mixed integer linear programming algorithm can be used to determine whether the constraints in $T$ have a solution. This step is also guaranteed to halt. Hence, Algorithm 1 always terminates.

**Theorem 2** Algorithm 1 is sound and complete in determining consistency of empirical programs, i.e. $\langle C, E \rangle$ is consistent iff Algorithm 1 declares that it is consistent.    □

**Example 6** Consider the empirical program:

$C$ :     $Monk\_seal(joe)$   $\leftarrow$
          $\neg Female(joe)$   $\leftarrow$
$E$ :   $[0.4, 0.45]\, Female(X)$   $\leftarrow$   $Monk\_seal(X)$

The following constraints are tested for satisfiability by Algorithm 1:

$$v_1 + v_2 \geq 1 \tag{1}$$
$$v_2 + v_4 \geq 1 \tag{2}$$
$$0.4 * (v_1 + v_2) \leq v_1 \leq 0.45 * (v_1 + v_2) \tag{3}$$
$$v_1 + v_2 + v_3 + v_4 \geq 1 \tag{4}$$

It is easy to verify that these constraints are satisfiable. For instance, one solution is: $v_1 = 4, v_2 = 6$ and $v_3, v_4$ equal to any non-negative integers. Thus, this program is consistent; for instance, the intepretation given in Example 3 is a model.    □

The above example highlights the major difference between the current framework for empirical probabilities and the earlier formalisms we proposed for subjective probabilities [13, 14]. In those formalisms, a clause corresponding in appearance to an emprical clause applies to *every* individual in the Herbrand domain. Thus, the subjective probability of $Female(joe)$ is simultaneously 0 due to $\neg Female(joe)$, and between 0.4 and 0.45 due to the clause in $E$. This gives rise to an inconsistency. Thus, this program is inconsistent in the formalisms in [13, 14].

## 4 Query Processing for Consistent Empirical Programs

### 4.1 Outline for Query Answering

A query to an empirical program is of the form: $Q \equiv F(d)$ which intuitively asks for the *conditional probability* of constant $d$ having (complex) property $F$, given that the program is true. As a preview, we first outline below a two-step procedure that can be used to answer this query; the procedure will be formalized in Section 4.3.

In the first step, the query answering procedure poses the query against the context $C$. If the context can deduce the truth or falsity of the query, then the procedure returns the answer 1 or 0 respectively, and the processing for the query is completed. Otherwise, the procedure moves to the second stage.

When no definite answer to the query can be *deduced*, the procedure then tries to *induce* the conditional probability by consulting the empirical clauses in $E$. More specifically, given the properties possessed by $d$, the procedure poses the query $F(X)$ to the empirical clauses. As we argued before, the $X$ here should not be taken as a quantifier; rather, it represents a generic *random* individual in the domain.

**Example 7** Consider the program discussed in Example 1. Suppose the query is $Q_1 \equiv Elephant(clyde)$.

Then, using the first and third clauses, it is easy to see that, given the context, the conditional probability of $Elephant(clyde)$ is 1. Now consider the query $Q_2 \equiv Grey(clyde)$. No definite answer for $Q_2$ can be deduced from the context. Thus, the query $Grey(X)$ is posed to the empirical clauses. The first two empirical clauses may be used to *induce* information about $clyde$ because $clyde$ is known, from the context, to be both an "Elephant" and a "Royal_elephant."    □

The query $Q_2$ in the above example underlines a major issue involved in such inductive answering – the choice of answers when more than one inductive answer is possible. The approach we take to resolve such conflicts is the one customarily used in statistical inferences – choose the one with the most specific reference class. In the following, before we formalize our query answering procedure, we first discuss how empirical programs can be compiled to facilitate the choice of reference classes and the corresponding empirical clauses.

### 4.2 Unfolding and Chaining of Empirical Programs

The unfolding procedure below computes the set of implications that are needed in query answering to determine the preferences of clauses.

**Algorithm 2** Let $C$ be the context of an empirical program.

1. Set $S$ to be the set of all complex predicate symbols in $\mathcal{L}$. Set $C'$ to be the set of all nonground clauses in $C$, i.e. $C' = \{L_0 \leftarrow L_1 \wedge \ldots \wedge L_n \mid L_0(X) \leftarrow L_1(X) \wedge \ldots \wedge L_n(X)$ is a clause in $C\}$.

2. Construct
   $impl(C) = \{F_1(X) \leftarrow F_2(X) \mid F_1, F_2 \in S$ and $\{\neg(F_1 \vee \neg F_2)\} \cup C'$ is unsatisfiable $\}$.    □

Recall that there are only finitely many predicate symbols in $\mathcal{L}$. Thus, in [15], we show that the set $S$ described in Algorithm 2 must be finite. Hence, the algorithm always terminates, yielding a finite set $impl(C)$.

Algorithm 2 is not the only optimization that can be carried out in compiling a program. Another one is *chaining* that can be computed by the following algorithm. The purpose of chaining clauses is to ensure that our query answering procedure (cf. Algorithm 4 below) is complete (cf. Theorem 4). In the sequel, the symbols $F_1, F_2, F_3$ and $F_4$ are assumed to be distinct complex predicate symbols.

**Algorithm 3** Let $EP = \langle C, E \rangle$ be an empirical program.

1. Use Algorithm 2 to compute the set $impl(C)$.

2. Set $T_0$ to $E$, and $i$ to 1.

3. Construct the set $S_1 = \{ [c_1, 1] F_1(X) \leftarrow F_3(X) \mid F_1(X) \leftarrow F_2(X)$ is a clause in $impl(C)$ and $[c_1, c_2] F_2(X) \leftarrow F_3(X)$ is a clause in $T_{i-1}\}$.

4. Construct the set $S_2 = \{ [1 - c_2, 1 - c_1] F_1(X) \leftarrow F_2(X) \mid [c_1, c_2] \neg F_1(X) \leftarrow F_2(X)$ is a clause in $T_{i-1}\}$.

5. Construct the set $S_3 = \{ [max\{0, c_1 + d_1 - 1\}, min\{c_2, d_2\}] (F_1 \wedge F_2)(X) \leftarrow F_3(X) \mid F_1 \neq F_2$ and $[c_1, c_2] F_1(X) \leftarrow F_3(X)$ and $[d_1, d_2] F_2(X) \leftarrow F_3(X)$ are clauses in $T_{i-1}\}$.

6. Construct the set $S_4 = \{ [max\{c_1, d_1\}, min\{1, c_2 + d_2\}] (F_1 \vee F_2)(X) \leftarrow F_3(X) \mid F_1 \neq F_2$ and $[c_1, c_2] F_1(X) \leftarrow F_3(X)$ and $[d_1, d_2] F_2(X) \leftarrow F_3(X)$ are clauses in $T_{i-1}\}$.

7. Construct the set $S_5 = \{ [0, min\{1, c_2 + d_2\}] (F_1 \wedge F_3)(X) \leftarrow F_2(X) \vee F_4(X) \mid F_1 \neq F_4, F_2 \neq F_4, F_2 \neq F_3$ and $[c_1, c_2] F_1(X) \leftarrow F_2(X)$ and $[d_1, d_2] F_3(X) \leftarrow F_4(X)$ are clauses in $T_{i-1}\}$.

8. Set $T_i = T_{i-1} \cup S_1 \cup S_2 \cup S_3 \cup S_4 \cup S_5$. If $T_i$ is the same as $T_{i-1}$, then set $comp(EP) = \langle impl(C), T_i \rangle$ and halt.

9. Otherwise, increment $i$ and go to Step 3.    □

Note that it is possible to have clauses of the form $F_1(X) \leftarrow F_2(X)$ and $F_2(X) \leftarrow F_1(X)$ in $impl(C)$ simultaneously. It is also possible to have empirical clauses $[c_1, c_2] F_1(X) \leftarrow F_2(X)$ and $[d_1, d_2] F_2(X) \leftarrow F_1(X)$ in $E$. Checking $T_i = T_{i-1}$ in Step 8 of Algorithm 3 is to prohibit cyclic chaining of clauses. In other words, the algorithm continues to the next iteration only if at least one new empirical clause is added in the current iteration. Since the number of predicate symbols in $\mathcal{L}$ is finite, there can only be a finite number of empirical clauses of the forms generated above. Thus, the above algorithm terminates after a finite number of iterations, yielding a finite program $comp(EP)$. Theorem 3 below shows that chaining does not change the meaning of an empirical program.

**Theorem 3** Let $EP = \langle C, E \rangle$ be an empirical program. Then $EP$ and $comp(EP)$ are logically equivalent.    □

### 4.3 A Procedure for Query Answering

Recall from Section 4.1 that inductive answering involves choosing the most specific reference class. However, in general, as reference classes may only follow a partial order in specificity, there may not be a unique most specific reference class; rather, there may only be several maximally specific reference classes. The following definition deals with this situation.

**Definition 7** Let $C$ be a context and $impl(C)$ be computed as shown above. Furthermore, let $S$ be a set of complex predicate symbols $\{F_1, \ldots, F_n\}$. A subset $T$ of $S$ is a *maximally preferred cluster* in $S$ based on $impl(C)$ if:

i) for any two elements $F_i, F_j \in T$, $F_i(X) \leftarrow F_j(X) \in impl(C)$ and vice versa, for some variable symbol $X$, and

ii) for all $F_j \in (S - T)$, $F_i(X) \leftarrow F_j(X) \notin impl(C)$, where $F_i \in T$. □

Basically, if two formulas $F_i, F_j$ are in the same cluster, then the context $C$ entails the formula $F_i(X) \leftrightarrow F_j(X)$, i.e. as far as the context $C$ is concerned, the formulas $F_i$ and $F_j$ are essentially equivalent. Intuitively, a maximally preferred cluster corresponds to a maximally specific reference class. This is formalized by Condition (ii) of the definition above.

**Algorithm 4** Let the inputs be a query $F(d)$ and the compiled version $comp(EP) = \langle C, E \rangle$ of an empirical program.

1. If $F(d)$ is a logical consequence of $C$, return 1 as the probability, and halt.

2. If $\neg F(d)$ is a logical consequence of $C$, return 0 as the probability, and halt.

3. (Inductive answering begins.) Otherwise, construct the set $S = \{ [c_1, c_2]F(X) \leftarrow F_1(X) \in E \mid F_1(d)$ is a logical consequence of $C\} \bigcup \{Cl \mid Cl \equiv [c_1, c_2]F(X) \leftarrow \in E\}$. Intuitively, $S$ contains all the applicable empirical clauses that either have a null body or require properties possessed by the individual $d$. If $S$ is empty, halt.

4. Otherwise, construct the set $S_1 = \{ F_1 \mid [c_1, c_2]F(X) \leftarrow F_1(X) \in S\}$. (For clauses in $S$ that have a null body, true is added to $S_1$.)

5. Compute the set of maximally preferred clusters in $S_1$ based on $impl(C)$. Let this set be $\{T_1, \ldots, T_m\}$.

6. For each $T_i$, $1 \leq i \leq m$, do the following. Let $T_i = \{F_1, \ldots, F_u\}$ for some $F_1, \ldots, F_u$. Return the range $\bigcap_{j=1}^{u} [c_1^j, c_2^j]$ as the probability range based on $F_1, \ldots, F_u$, where the clauses $[c_1^j, c_2^j] F(X) \leftarrow F_j(X)$ for all $1 \leq j \leq u$ are all in $S$. □

Note that Steps 1, 2 and 3 of Algorithm 4 involve checking to see whether a ground atom of the form $F(d)$ is true or false in all the models of the given context $C$. Such testing can be implemented by a standard unsatisfiability checker, such as one based on resolution, or a mixed integer linear programming algorithm such as the one in [4].

**Example 8** Suppose the query $\neg White(clyde)$ is posed against the program considered in Example 1. It is easy to see that there are models of the context $C$ in which $\neg White(clyde)$ is true as well as models of $C$ in which $\neg White(clyde)$ is false. Thus, inductive answering in Step 3 of Algorithm 4 is needed. Since it is obvious that both $Royal\_elephant(clyde)$ and $Elephant(clyde)$ are true in all models of $C$, the two clauses below (which are generated when the program is compiled by Algorithm 3) are added to the set $S$ in Step 3:

$$[0, 0.1] \neg White(X) \leftarrow Royal\_elephant(X)$$
$$[0, 0.95] \neg White(X) \leftarrow Elephant(X)$$

The set $S_1$ constructed in Step 4 of Algorithm 4 is $\{Elephant, Royal\_elephant\}$. Since the clause $Elephant(X) \leftarrow Royal\_elephant(X)$ is in $C$, it is obvious that there is only one maximally preferred cluster – namely the singleton set $\{Royal\_elephant\}$. Hence, the answer of the query is the range $[0, 0.1]$, and the range $[0, 0.95]$ based on $Elephant$ is not returned as an answer. □

Given a compiled empirical program $comp(EP)$ and a query $F(d)$, we use the notation $proof(EP, F, d)$ to denote the set of all ranges [2] obtained by applying Algorithm 4 to $comp(EP)$ and $F(d)$. We also use the notation $consq(EP, F, d)$ to denote the set of all ranges satisfied by all the models of $EP$ for the query $F(d)$.

**Theorem 4** Let $EP = \langle C, E \rangle$ be an empirical program and $F(d)$ be a query. Then Algorithm 4 is sound and complete, i.e. $proof(EP, F, d) = consq(EP, F, d)$. □

## 5  Related Work

Our framework was inspired, in part, by Bacchus' framework that extends full first order logic with empirical probability statements. He develops a sound and complete proof procedure for consistent theories, but does not provide a consistency check mechanism. We provide explicit mechanisms for determining the consistency of empirical programs. These mechanisms are based upon mixed integer linear programming techniques and hence, may be implemented on top of standard integer programming packages. They are guaranteed to terminate. Both Bacchus [2] and our framework use the concept of inductive answering. Furthermore, we provide compilation techniques that facilitate processing of queries that require inductive (and deductive) answering.

The integration of logic and probability theory has been the subject of numerous studies [7, 10, 16, 18]. Fagin, Halpern and Meggido [7] have proposed a model-theoretic basis for reasoning about systems of linear inequalities of probabilities. Their proposal is based on the possible-world approach, and hence is different from the empirical approach considered here. Kavvadias and Papadimitriou [10] have studied the consistency of probabilistic logic programs using linear programming. However, their study is based on Nilsson's possible worlds semantics, i.e. the linear programs specify constraints on the probability distributions over the set of possible worlds (which is identical to the set of Herbrand interpretations). On the other hand, the linear programs in our framework specify constraints over the domain of an interpretation, which may or may not be an Herbrand interpretation.

---

[2]The probabilities 1 and 0 can be represented by the ranges $[1,1]$ and $[0,0]$ respectively.

# 6   Conclusions

In this paper, we develop a model-theoretic foundation for logic programming that supports empirical probabilities and allows inductive probabilistic reasoning to be performed. To capture the intuitions behind empirical probabilities, our model theory also considers non-Herbrand interpretations. Furthermore, we develop a totally correct algorithm for verifying the consistency of monadic deductive databases. This algorithm can be implemented using mixed integer linear programming algorithms. Finally, for dealing with queries to such databases, we develop a sound and complete procedure that uses deductive and inductive answering. We also devise compilation techniques to facilitate this procedure.

### Acknowledgements

This research was partially sponsored by the National Science Foundation under Grant IRI-91-09755.